\documentclass[conference]{IEEEtran}
\IEEEoverridecommandlockouts
\usepackage{cite}
\usepackage{url}
\usepackage{amsmath,amssymb,amsfonts}
\usepackage{graphicx}
\usepackage{textcomp}
\usepackage{xcolor}
\usepackage{tikz}
\usetikzlibrary{arrows.meta,positioning}
\usepackage{booktabs}
\usepackage[pdfauthor={Mansur Ozaman},pdftitle={Evaluating Deep Learning and Traditional Approaches Used in Source Camera Identification},pdfsubject={},pdfkeywords={},pdfcreator={},pdfproducer={},hidelinks]{hyperref}

\def\BibTeX{{\rm B\kern-.05em{\sc i\kern-.025em b}\kern-.08em
    T\kern-.1667em\lower.7ex\hbox{E}\kern-.125emX}}

\begin{document}

\title{Evaluating Deep Learning and Traditional Approaches Used in Source Camera Identification}

\author{
\IEEEauthorblockN{Mansur Ozaman}
\IEEEauthorblockA{Department of Computer Science\\
School of Computing and Artificial Intelligence\\
Nazarbayev University\\
Astana, Kazakhstan\\
mansur.ozaman@nu.edu.kz}
}

\maketitle

\begin{abstract}
Identifying the device that captured a digital image is a central task
in image forensics, made more urgent by pervasive smartphone
photography and the rise of AI-generated imagery. This paper compares
three source camera identification (SCI) approaches -- Photo Response
Non-Uniformity (PRNU) analysis, JPEG compression artifact analysis,
and a convolutional neural network (CNN) -- on images from four
smartphone devices (Xiaomi Redmi Note~10S, Samsung Galaxy S24,
iPhone~13 Pro Max, and iPhone~17 Pro). For each method, discriminative
features are extracted (DCT coefficient statistics for JPEG artifacts,
denoised noise residuals for PRNU, and learned convolutional features
for the CNN) and classified with an SVM (JPEG, PRNU) or trained
end-to-end (CNN), using an identical fixed, stratified 70/30
train--test split for direct comparability. The DCT/JPEG-artifact-inspired
classifier achieves the highest accuracy (0.90), followed by the
PRNU-inspired method (0.71), while the CNN baseline reaches only 0.29.
Under this experimental configuration, the DCT/statistical and
PRNU-inspired approaches outperform the evaluated CNN baseline. The
results motivate evaluation on larger, format-balanced datasets and
investigation of hybrid or learned approaches under more robust
validation protocols.
\end{abstract}


\section{Introduction}

According to the World Bank Group's Global Findex Digital
Connectivity Tracker (2025), 68\% of adults worldwide own a
smartphone \cite{worldbank2025}. Determining the source device of a
digital image has become a crucial task given such widespread
smartphone usage: images can be tampered with or posted in misleading
contexts, and source camera identification (SCI) helps restore their
authenticity and credibility. SCI is also important for multimedia
security and intellectual property protection, and, as AI
image-generating models advance, for maintaining the integrity of
digital media more broadly.

Traditional SCI techniques rely on sensor characteristics. In natural
images the dominant part of the pattern noise is the photo-response
non-uniformity noise (PRNU), caused primarily by the differing
sensitivity of pixels to light from inhomogeneity of silicon wafers
and manufacturing imperfections \cite{lukas2006digital}; this study
focuses on PRNU for its ubiquity. A second traditional feature class
derives from JPEG compression artifacts: JPEG's lossy compression
introduces blocking and quantization patterns that can identify an
image's source, particularly when metadata is stripped or images are
heavily compressed \cite{kwon2022jpeg}. More recently, as
metadata-stripping social-media sharing and filter-based editing
distort these patterns, convolutional neural network (CNN)-based
techniques have emerged, often regarded as superior for their ability
to learn features directly from the data.

This paper presents a comparative analysis of the three
methods -- PRNU, JPEG artifact analysis, and CNN -- on a shared
smartphone-image dataset, evaluated under an identical train--test
protocol and compared on accuracy.

\section{Literature Review}
\label{sec:litreview}

Source camera identification has shifted increasingly from purely
physical sensor-noise models toward learned, data-driven
representations. Mandelli et al.~\cite{mandelli2020} showed that
patch-level convolutional features can approach PRNU-based accuracy at
far less computational cost, while Sameer and Naskar~\cite{sameer2020}
addressed the difficulty of collecting large labeled datasets with a
Siamese network for limited-labels classification. Wang et
al.~\cite{wang2021csicnn} extended CNN-based identification to
IoT/social-media contexts where compression and re-sharing degrade
sensor-noise fingerprints, and Chen and Stamm~\cite{chen2021demosaic}
strengthened camera-\emph{model} identification using demosaicing
residuals that generalize better across devices sharing similar
sensors.

Later work focused on datasets and architectures suited to
limited-data regimes close to the present setting. Bernacki and
Scherer~\cite{bernacki2023imagine} introduced the IMAGINE benchmark of
CMOS-sensor devices, and Wang et al.~\cite{wang2023mdmcps} proposed a
metric-learning approach (MDM-CPS) for few-shot source camera
identification. Akbari et al.~\cite{akbari2024hierarchical} and Huan
et al.~\cite{huan2024convnext} refined deep architectures (fusion-layer
networks; a dual-path ConvNeXt with patch selection) for model-level
identification, while Nwokeji et al.~\cite{nwokeji2024survey} surveyed
hardware- versus software/deep-learning-based techniques and found
that no single family dominates across all conditions. More recently,
Elharrouss et al.~\cite{elharrouss2025pdcvit} proposed the
transformer-based PDC-ViT, and Croitoru et
al.~\cite{croitoru2025prnubench} released PRNU-Bench, a large
``in-the-wild'' PRNU benchmark addressing the reliance of prior PRNU
evaluations on outdated or overly clean data. Most directly relevant
here, Li and Law~\cite{li2025recentadvances} survey deep-learning SCI
architectures and report that, while these models achieve strong
camera-\emph{model}-level classification, their \emph{device}-level
performance remains substantially weaker than traditional PRNU-based
fingerprinting, and that modern in-camera processing degrades both
classical and learned fingerprints.

This body of work suggests two trends relevant to the present study.
First, deep-learning SCI methods are typically developed on large or
purpose-built datasets, and performance can degrade sharply with only
a handful of images per device, which may help explain the
comparatively low CNN accuracy observed here. Second, several recent
surveys and
benchmarks~\cite{nwokeji2024survey,croitoru2025prnubench,li2025recentadvances}
still find engineered, interpretable features -- PRNU and
compression-based statistics -- highly competitive at the device
level, motivating the three-way comparison performed in this paper.

\section{Methods}

\subsection{Overall Pipeline}

Figure~\ref{fig:pipeline} summarizes the overall processing pipeline
used for all three methods: images are first loaded from the raw
dataset, then passed through a method-specific preprocessing stage
(grayscale conversion and DCT block extraction for JPEG; residual
denoising for PRNU; resizing and normalization for the CNN), before
being classified by the corresponding model to produce a predicted
device label.

\begin{figure}[t]
\centering
\begin{tikzpicture}[
  node distance=4mm,
  box/.style={draw, rounded corners, minimum width=4.0cm, minimum height=7mm,
              align=center, font=\footnotesize, inner sep=2pt},
  arrow/.style={-{Latex[length=1.8mm]}, thick}
]
\node[box] (data) {Raw dataset (4 devices)};
\node[box, below=of data] (prep) {Preprocessing};
\node[box, below=of prep] (pipe) {Feature extraction\\+ classifier\\(JPEG / PRNU / CNN)};
\node[box, below=of pipe] (out) {Predicted device};

\draw[arrow] (data) -- (prep);
\draw[arrow] (prep) -- (pipe);
\draw[arrow] (pipe) -- (out);
\end{tikzpicture}
\caption{Overall methods pipeline: dataset, preprocessing, and the
feature extraction / classification stage shared by all three SCI
methods.}
\label{fig:pipeline}
\end{figure}

\subsection{Dataset and Preprocessing}

I evaluate three families of source camera identification methods on
a dataset consisting of four smartphone devices: Xiaomi Redmi Note
10S, Samsung Galaxy S24, iPhone 13 Pro Max, and iPhone 17 Pro. The
dataset comprises 68 images in total, 17 per device, captured at
default camera settings and transferred via direct cable connection to
avoid the recompression and metadata stripping introduced by
messaging and cloud-sharing services. The data directories are
structured as \texttt{data/redminote10s/}, \texttt{data/samsungs24/},
\texttt{data/iphone13promax/}, and \texttt{data/iphone17pro/}. Redmi
and Samsung images are stored as JPEG files, while both iPhone models
provide images in HEIC (HEIF) format. Each device is a single physical
unit, and the four devices also correspond to four distinct camera
models; consequently, this experiment evaluates classification among
these four device/model classes and does not establish individual
same-model device attribution (i.e., distinguishing between two
different physical units of the same model), since no second unit of
any model was available. The dataset used in this work is publicly
available at \cite{anon2025dataset}.

All images are loaded using Python and decoded to RGB using Pillow
\cite{pillow} and pillow-heif \cite{pillowheif}, and converted
to NumPy arrays \cite{harris2020array}.

For the JPEG artifact and PRNU-based pipelines, images are converted
to grayscale using the standard ITU-R BT.601 luma transform. For the
CNN classifier, the RGB images are resized to $128 \times 128$ pixels
using bilinear interpolation and normalized to the $[0,1]$ range by
dividing by 255.

For each method I employ the same stratified 70/30 train-test split
(stratified by device class) with a fixed random seed, to make the
results directly comparable across the three methods. Given the
dataset size, this yields approximately 12 training and 5--6 test
images per device class; the resulting small test-set sizes are a
limitation of this study and are discussed in Section~\ref{sec:limitations}.
The code used to process and analyze the images is publicly available
\cite{anon2025code}.

\subsection{DCT/JPEG-Artifact-Inspired Features}
\label{sec:jpeg-methods}

JPEG compression is block-based and operates by applying the
two-dimensional discrete cosine transform (DCT) to non-overlapping
$8 \times 8$ blocks, followed by scalar quantization and entropy
coding \cite{wallace1992jpeg}. Different camera models and firmware
typically use different quantization tables and pre-processing steps,
producing distinct distributions of DCT coefficients; prior work has
shown that such coefficient statistics reveal JPEG compression history
and quantization parameters \cite{fan2003identification}. This work
exploits that quantization-related variation as a discriminative
feature for device attribution.

Let $I$ denote a grayscale image of size $H \times W$. To ensure an
integer number of blocks, I crop the image so that both dimensions are
multiples of 8:
\begin{equation}
\begin{split}
H' &= H - (H \bmod 8), \\
W' &= W - (W \bmod 8),
\end{split}
\end{equation}
and retain only the top-left $H' \times W'$ region. The image is then
partitioned into $M = (H'/8) \cdot (W'/8)$ non-overlapping blocks
$\{b_m\}_{m=1}^{M}$, each of size $8 \times 8$.

For each block $b_m$, I compute the 2D DCT according to
\begin{equation}
\begin{split}
C_m(u,v) = {}&\alpha(u)\alpha(v) \sum_{x=0}^{7}\sum_{y=0}^{7}
b_m(x,y) \\
&\times\cos\!\left[\frac{(2x{+}1)u\pi}{16}\right]
\cos\!\left[\frac{(2y{+}1)v\pi}{16}\right],
\end{split}
\label{eq:dct}
\end{equation}
for $u,v = 0,\dots,7$, with $\alpha(0) = 1/\sqrt{8}$ and
$\alpha(k) = 1/2$ for $k = 1,\dots,7$.

The DC coefficient $(u,v)=(0,0)$ together with the remaining
coefficients in its row and column ($u=0$ or $v=0$) are discarded,
retaining the $7\times7$ block of higher-order AC coefficients with
$u,v \in \{1,\dots,7\}$, which are more directly shaped by
quantization. For each retained position, I compute the empirical mean
and variance of the coefficient over all blocks:
\begin{equation}
\begin{split}
\mu_{u,v} &= \frac{1}{M}\sum_{m=1}^{M} C_m(u,v), \\
\sigma^2_{u,v} &= \frac{1}{M-1}\sum_{m=1}^{M} \big(C_m(u,v) - \mu_{u,v}\big)^2 .
\end{split}
\label{eq:jpeg_stats}
\end{equation}

The JPEG artifact feature vector $f$ is formed by concatenating all
49 means and 49 variances of these retained AC coefficients:
\begin{equation}
\begin{split}
f = \Big[\;&\{\mu_{u,v}\}_{u,v=1,\dots,7}, \\
&\{\sigma^2_{u,v}\}_{u,v=1,\dots,7}\Big]^\top \in \mathbb{R}^{98}.
\end{split}
\label{eq:jpeg_feat}
\end{equation}

Before classification, $f$ is reduced to 20 dimensions with principal
component analysis (PCA, fit once on the full feature set prior to the
train--test split) and each resulting dimension is standardized to
zero mean and unit variance. I then use a support vector machine (SVM)
with a radial basis function (RBF) kernel \cite{cortes1995support} as
the classifier, implemented using scikit-learn \cite{pedregosa2011scikit}.
Even for HEIC images, the same pipeline is applied to the decoded
grayscale image, so that all devices are processed uniformly in the
spatial domain.

Two clarifications follow from this. First, the resulting features are
block-wise DCT coefficient statistics of the \emph{decoded} image, not
the original quantized JPEG coefficients read from a compressed
bitstream; for the two iPhone devices, which supply HEIC images, these
statistics are computed after HEIC decoding rather than from a native
JPEG encoder, so the method is more accurately described as a
DCT/statistical feature method than as literal JPEG-coefficient
analysis. Second, because Redmi and Samsung images are natively JPEG
while both iPhone models are natively HEIC, native image format is
confounded with device identity in this dataset: the strong separation
achieved by this method may partly reflect format- and
firmware-specific processing differences rather than sensor
characteristics alone. This comparison should therefore be read as
performance under the native-format conditions of this particular
dataset (Section~\ref{sec:limitations}), not as isolating a
device-intrinsic signal.

\subsection{PRNU-Inspired Noise-Residual Features}

Photo Response Non-Uniformity (PRNU) originates from slight variations
in pixel sensitivities on the image sensor. It can be modeled as
\cite{lukas2006digital}
\begin{equation}
I = I_0(1+K) + \Theta,
\end{equation}
where $I$ is the observed image, $I_0$ is the ideal noise-free image,
$K$ is the PRNU pattern, and $\Theta$ summarizes other noise sources.
PRNU can be obtained from a noise residual and used as a
device-specific fingerprint.

Each image is first resized to $512 \times 512$ pixels, then converted
to grayscale, cast to floating point, and normalized to $[0,1]$. A
high-frequency residual is extracted by subtracting a smoothed version
of the image:
\begin{equation}
R = I - G_{\sigma_1} * I,
\label{eq:residual}
\end{equation}
where $G_{\sigma_1}$ is a Gaussian kernel with standard deviation
$\sigma_1$ (set to 1 in this implementation) and $*$ denotes
convolution.

To suppress non-PRNU components, I estimate the local mean and
variance of the residual using another Gaussian filter $G_w$ (scale
$w=5$) \cite{lukas2006digital}:
\begin{equation}
\begin{split}
\mu(x) &= G_w * R(x), \\
\sigma^2(x) &= G_w * R^2(x) - \mu^2(x),
\end{split}
\end{equation}
and form a Wiener-style estimate
\begin{equation}
\begin{split}
\hat{R}(x) = \bigl(R(x)-\mu(x)\bigr) \times {}&\\
&\frac{\sigma^2(x)}{\sigma^2(x) + \sigma_n^2},
\end{split}
\label{eq:wiener}
\end{equation}
with $\sigma_n^2 = 0.01$ a small constant. The PRNU-like pattern is
then
\begin{equation}
K(x) = R(x) - \hat{R}(x).
\label{eq:prnu_pattern}
\end{equation}

The pattern $K$ is flattened into a one-dimensional vector
$g = \mathrm{vec}(K)$, and the first 50{,}000 entries of $g$ are
retained as the final feature vector, for computational convenience.
As in the JPEG case, the features are standardized and fed into a
linear SVM \cite{pedregosa2011scikit}, following the general approach
used in large-scale PRNU-based camera identification
\cite{goljan2009large}.

This procedure differs from conventional PRNU-based identification.
Conventional PRNU-based identification
\cite{lukas2006digital,goljan2009large} estimates a device-specific
reference fingerprint by averaging noise residuals over many images
from the same device, and then attributes a query image by correlating
its residual against each candidate fingerprint. The pipeline
evaluated here instead extracts a single-image noise residual $K$ per
photo (Eqs.~\ref{eq:wiener}--\ref{eq:prnu_pattern}), retains a fixed
subset of its entries as a feature vector, and classifies that vector
directly with a linear SVM -- no multi-image fingerprint is estimated
and no correlation-based matching is performed. This is therefore best
described as a \emph{PRNU-inspired, noise-residual feature} method
rather than full PRNU fingerprint estimation and matching, and its
results should be interpreted accordingly; it is referred to as
``PRNU-inspired'' throughout the remainder of the paper.

\subsection{CNN-Based Classifier}
\label{sec:cnn-methods}

The third approach uses a convolutional neural network trained
end-to-end on RGB images to directly learn features for camera
identification, in the spirit of Bondi et al.~\cite{bondi2017first}. Each
input image is resized to $128 \times 128 \times 3$ and scaled to the
$[0,1]$ range. The network architecture is implemented using
TensorFlow \cite{tensorflow2015} and Keras \cite{chollet2015keras}:

\begin{enumerate}
\item A convolutional layer with 32 filters of size $3\times3$,
followed by ReLU activation and $2\times2$ max pooling.
\item A convolutional layer with 64 filters of size $3\times3$,
followed by ReLU activation and $2\times2$ max pooling.
\item A convolutional layer with 128 filters of size $3\times3$,
followed by ReLU activation.
\item A flatten layer.
\item A fully connected layer with 64 neurons and ReLU activation.
\item An output fully connected layer with 4 neurons (one per device
class) and softmax activation.
\end{enumerate}

For an input image $X_i$, the final layer produces logits $z_{i,k}$
for each class $k \in \{1,\dots,4\}$. The corresponding posterior
probabilities are given by the softmax function
\begin{equation}
p_{i,k} = \frac{\exp(z_{i,k})}{\sum_{j=1}^{4}\exp(z_{i,j})}.
\label{eq:softmax}
\end{equation}

Training uses the categorical cross-entropy loss
\begin{equation}
\mathcal{L} = -\frac{1}{N}\sum_{i=1}^{N}\sum_{k=1}^{4} y_{i,k}\log p_{i,k},
\label{eq:cross_entropy}
\end{equation}
where $N$ is the number of training samples and $y_{i,k}$ is the
one-hot encoding of the ground-truth label of image $i$. The network
is optimized using the Adam optimizer \cite{kingma2014adam} for five
epochs with mini-batches of size 8.

\subsection{Evaluation Metrics}

For each method, classification performance is measured using the
overall accuracy on the test set:
\begin{equation}
\mathrm{Accuracy} = \frac{1}{N_{\mathrm{test}}}\sum_{i=1}^{N_{\mathrm{test}}} \mathbb{1}[\hat{y}_i = y_i],
\label{eq:accuracy}
\end{equation}
where $N_{\mathrm{test}}$ is the number of test samples, $y_i$ is the
ground-truth device label, $\hat{y}_i$ is the predicted label, and
$\mathbb{1}[\cdot]$ is the indicator function. Normalized confusion
matrices are also computed for each method using Matplotlib
\cite{hunter2007matplotlib} and Seaborn \cite{waskom2021seaborn}; each row
corresponds to the true device and each column to the predicted
device, with rows summing to one, so that the diagonal entries are
exactly the per-class recalls. Macro-averaged recall, reported in
Table~\ref{tab:prf}, is computed as the unweighted mean of these
diagonal entries. Macro-averaged precision (also macro-averaged over
the four classes, for consistency with recall) additionally requires
the non-normalized, count-based confusion matrix. Because the
train--test split is deterministic (fixed \texttt{random\_state},
stratified by class), the exact per-class test-set sizes were
recovered by re-running the published splitting procedure on the
class labels (5, 5, 6, and 5 test images for
\texttt{iphone13promax}, \texttt{iphone17pro}, \texttt{redminote10s},
and \texttt{samsungs24}, respectively; 21 test images in total); these
counts convert the published normalized confusion matrices into exact
integer prediction counts, from which precision is computed. As a
consistency check, the resulting accuracy from these reconstructed
counts (19/21, 15/21, and 6/21 for JPEG, PRNU, and CNN) matches the
originally reported 0.90, 0.71, and 0.29. Where a class receives no
predicted positives (as for the CNN on two devices; see
Section~\ref{sec:limitations}), precision for that class is undefined;
for computation of the reported macro-precision value, these cases are
assigned a precision of zero, consistent with the zero-division
handling used in the metric calculation.

\section{Results}

\subsection{Overall Classification Accuracy}

\begin{figure}[t]
\centering
\includegraphics[width=0.9\columnwidth]{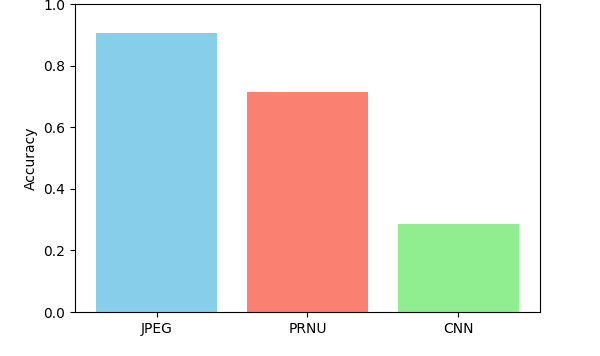}
\caption{Overall accuracy comparison of the JPEG-based, PRNU-inspired,
and CNN-based classification methods.}
\label{fig:accuracy}
\end{figure}

Figure~\ref{fig:accuracy} summarizes the overall accuracy of the three
methods: $\mathrm{Acc_{JPEG}} = 0.90$, $\mathrm{Acc_{PRNU}} = 0.71$,
and $\mathrm{Acc_{CNN}} = 0.29$, a clear separation between the
traditional feature approaches and the end-to-end CNN in this
experimental setting. The CNN's accuracy, while lowest, still reflects
non-random patterns and serves as a baseline for learned-feature
comparison. Table~\ref{tab:prf} reports accuracy, macro-averaged
precision and recall, and per-device recall for all three methods;
macro recall and accuracy are close for JPEG and PRNU but diverge
slightly for the CNN (0.30 vs.\ 0.29), reflecting the slightly unequal
per-class test-set sizes noted in Section~\ref{sec:limitations}.

\begin{table*}[t]
\centering
\caption{Classification performance by method, including overall and per-device metrics.}
\label{tab:prf}
\begin{tabular}{lccc cccc}
\toprule
& & & & \multicolumn{4}{c}{\textbf{Per-Device Recall}} \\
\cmidrule(lr){5-8}
\textbf{Method} & \textbf{Accuracy} & \textbf{Macro Prec.} & \textbf{Macro Recall} &
\textbf{iPhone 13 PM} & \textbf{iPhone 17 Pro} & \textbf{Redmi 10S} & \textbf{Galaxy S24} \\
\midrule
DCT/JPEG-artifact-inspired & 0.90 & 0.93 & 0.90 & 1.00 & 0.60 & 1.00 & 1.00 \\
PRNU-inspired (residual)   & 0.71 & 0.80 & 0.70 & 0.20 & 0.60 & 1.00 & 1.00 \\
CNN (learned features)     & 0.29 & 0.19 & 0.30 & 1.00 & 0.00 & 0.00 & 0.20 \\
\bottomrule
\end{tabular}
\end{table*}

\subsection{Per-Device Behaviour and Comparison of the Methods}

\begin{figure}[t]
\centering
\includegraphics[width=0.9\columnwidth]{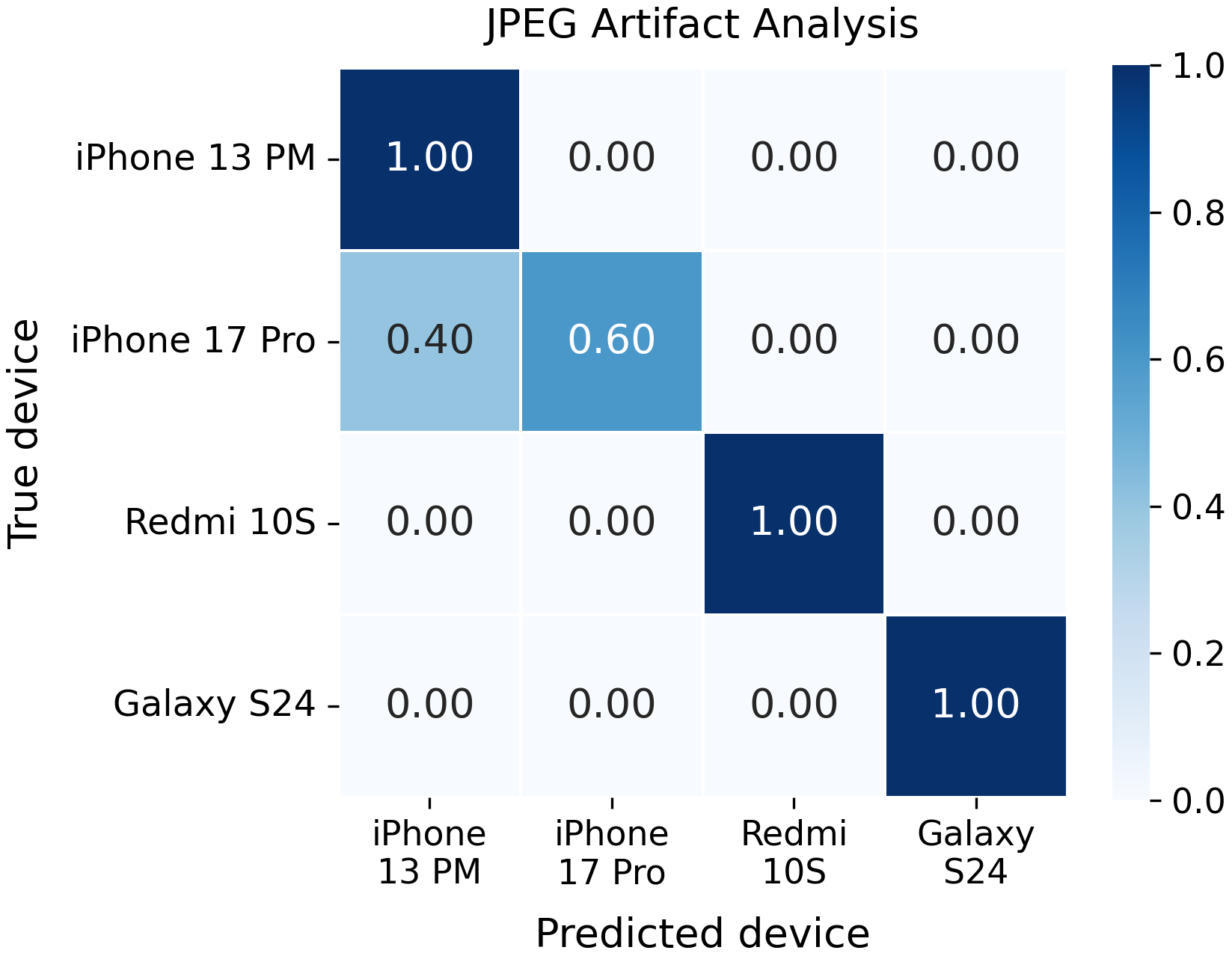}
\caption{Confusion matrix for the JPEG artifact-based classifier.}
\label{fig:cm-jpeg}
\end{figure}

\begin{figure}[t]
\centering
\includegraphics[width=0.9\columnwidth]{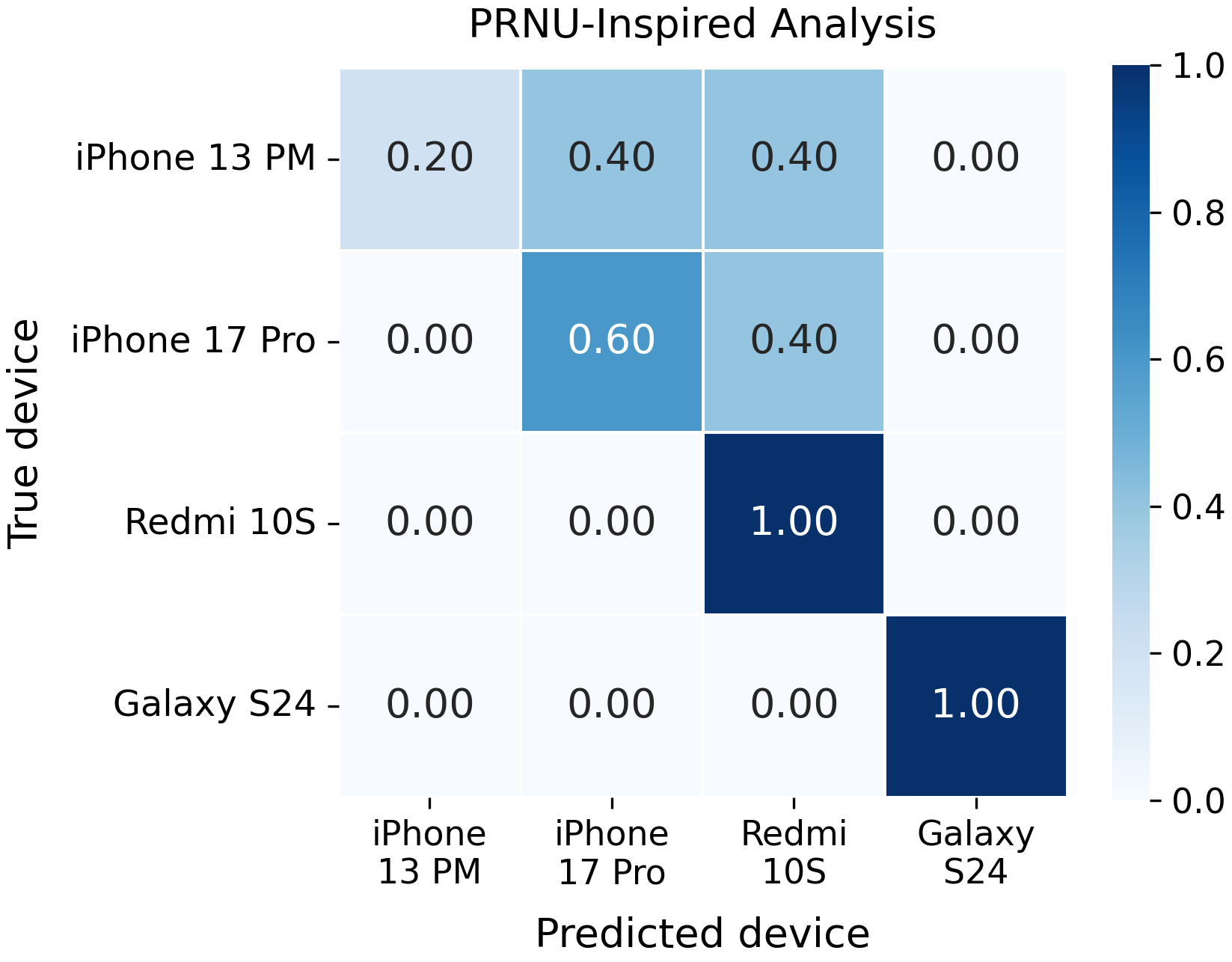}
\caption{Confusion matrix for the PRNU-inspired classifier.}
\label{fig:cm-prnu}
\end{figure}

\begin{figure}[t]
\centering
\includegraphics[width=0.9\columnwidth]{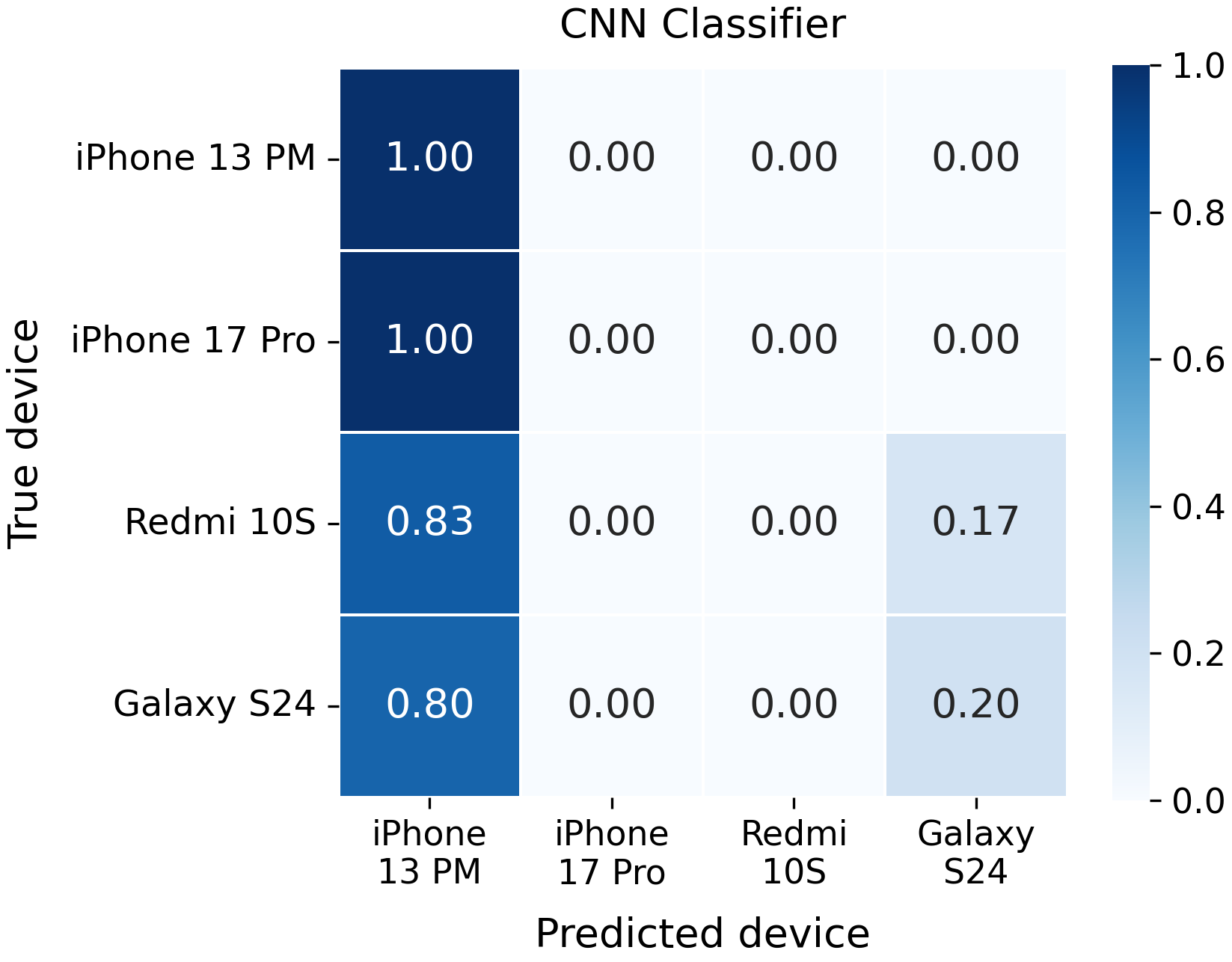}
\caption{Confusion matrix for the CNN classifier.}
\label{fig:cm-cnn}
\end{figure}

Table~\ref{tab:prf} summarizes per-device recall; Figures~\ref{fig:cm-jpeg}--\ref{fig:cm-cnn}
show the underlying normalized confusion matrices (rows are the true
device, columns the predicted device) and reveal the error patterns
behind those numbers. Both the JPEG (Figure~\ref{fig:cm-jpeg}) and
PRNU-inspired (Figure~\ref{fig:cm-prnu}) classifiers separate the two
Android devices perfectly and make most of their errors between the
two iPhone models: JPEG assigns 40\% of \texttt{iphone17pro} images to
\texttt{iphone13promax}, while PRNU spreads \texttt{iphone13promax}
predictions across all three non-\texttt{samsungs24} classes. The CNN
(Figure~\ref{fig:cm-cnn}) behaves qualitatively differently, collapsing
most predictions onto \texttt{iphone13promax} -- including all
\texttt{iphone17pro} images and the majority of Android images
(0.83 and 0.80 for \texttt{redminote10s} and \texttt{samsungs24}) --
which explains both its low accuracy and its highly asymmetric
per-device recall in Table~\ref{tab:prf}.

Under the experimental conditions evaluated here, the JPEG classifier
gives the highest and most balanced performance across all four
devices; the PRNU-inspired classifier follows closely, matching JPEG
on the Android devices but with less consistent iPhone results; and
the CNN, used as an end-to-end learned baseline, remains well below
both traditional approaches, largely favoring a single dominant class.

\subsection{Limitations}
\label{sec:limitations}

Several limitations should be considered when interpreting these
results. The dataset is small: 68 images across four devices (17 per
device) yield only 5--6 test images per class after the 70/30 split,
so reported accuracies rest on very few test examples and single-point
estimates should not be read as precise population accuracies; a small
change in a handful of test predictions would noticeably shift the
reported numbers. Results are also reported for a single, fixed
train--test split rather than averaged over repeated splits or
$k$-fold cross-validation, so no variance or confidence interval is
available.

The CNN is a three-convolutional-layer network
(Section~\ref{sec:cnn-methods}) trained from scratch, with no
pretraining or data augmentation, for five epochs on roughly 12 images
per class; the limited training data, this specific architecture and
training configuration, and the small evaluation set may each have
contributed to its low accuracy, though their individual contributions
cannot be disentangled with the present experiment. Its predictions
collapse largely onto a single class (Table~\ref{tab:prf},
Figure~\ref{fig:cm-cnn}), leaving precision undefined for two of the
four devices; this result should not be read as a general statement
about CNN-based SCI.

Native image format is also confounded with device identity (two
Android devices producing JPEG, two iPhone models producing HEIC), so
the strong performance of the DCT/JPEG-artifact-inspired method may
partly reflect format and firmware differences rather than sensor
characteristics alone. In addition, each camera model is represented
by a single physical unit, so the experiment demonstrates
classification among these four device/model classes and cannot speak
to identification between two units of the same model. Finally, in the
DCT/JPEG-artifact-inspired pipeline, PCA is fit on the full
98-dimensional feature set before the train--test split
(Section~\ref{sec:jpeg-methods}), so the learned PCA transformation is
exposed to held-out test images during fitting; this is a form of
preprocessing leakage, and the reported 0.90 accuracy for this method
should therefore be interpreted as an optimistic estimate of
generalization performance rather than an unbiased one. This
experiment does not measure how large the resulting bias is. Future
evaluation should fit PCA exclusively on the training partition and
apply that fitted transformation to the held-out test set.
Collectively, these limitations restrict how confidently the observed
ranking among
the three methods can be generalized beyond the present dataset and
split; they do not invalidate the ranking as a description of what was
observed here.

\section{Conclusion}

This work presented a comparative study of three source camera
identification methods on a multi-device smartphone dataset: JPEG
artifact-based features, PRNU-inspired noise-residual features, and a
convolutional neural network classifier, achieving 0.90, 0.71, and
0.29 accuracy respectively (Table~\ref{tab:prf}). The CNN's behaviour
was distinct from the other two, concentrating its predictions on a
single dominant class rather than distributing errors across devices.

Within the scope of this small, four-device dataset, the observed
ranking indicates that the traditional, interpretable feature-based
methods outperformed the CNN trained from scratch under this
configuration; as discussed in Section~\ref{sec:limitations}, this gap
is consistent with the limited-data regime rather than a general claim
about CNN-based SCI. The ranking is further qualified by the confound
between native image format and device identity, and by PCA being fit
before the train--test split in the DCT/JPEG-artifact-inspired
pipeline, which may optimistically bias its reported accuracy;
stronger validation on a larger, format-balanced dataset with
cross-validated splits is needed before generalizing these results.
Future work can also investigate hybrid approaches that combine
PRNU-inspired and JPEG-artifact-inspired features with learned
representations, as well as transfer learning or pretraining
strategies to make CNN-based SCI competitive under limited data.


\end{document}